\documentclass[conference, 10pt]{IEEEtran}
\IEEEoverridecommandlockouts
\usepackage{cite}
\usepackage{amsmath,amssymb,amsfonts}
\usepackage{graphicx}
\usepackage{tikz}

\usepackage{nomencl}
\usepackage{amsthm}
\usepackage{diagbox}
\usepackage[colorlinks,
citecolor=purple,
urlcolor=purple,
linkcolor=purple,
bookmarks=false,
hypertexnames=true]{hyperref}
\usepackage{soul}
\usepackage{multirow}
\usepackage{textcomp}
\usepackage{enumitem}
\usepackage{xcolor}
\usepackage[normalem]{ulem}
\usepackage{caption}
\usepackage{subcaption}
\usepackage{algpseudocode}
\usepackage[ruled,vlined,linesnumbered]{algorithm2e}
\usepackage{algorithmicx}
\usepackage{flushend}

\makenomenclature

\newcommand{\mr}[1]{{\textcolor{blue}{}}}
\newcommand{\lc}[1]{{\textcolor{magenta}{}}}
\newcommand{\jd}[1]{{\textcolor{green}{}}}
\newcommand{\zz}[1]{{\textcolor{orange}{}}}

\newcommand{\algo}[1]{%
{\fontfamily{qcr}\selectfont TabVFL}\xspace#1%
}
\newcommand{\stepnum}[1]{%
\begin{tikzpicture}
    \node[draw, circle, fill=black, text=white, inner sep=0.5pt, minimum size=1mm] {#1};
\end{tikzpicture}%
}
\newcommand{\codedtext}[1]{%
{\fontfamily{qcr}\selectfont #1}%
}

\def\BibTeX{{\rm B\kern-.05em{\sc i\kern-.025em b}\kern-.08em
    T\kern-.1667em\lower.7ex\hbox{E}\kern-.125emX}}

\newcommand\blfootnote[1]{%
  \begingroup
  \renewcommand\thefootnote{}\footnote{#1}%
  \addtocounter{footnote}{-1}%
  \endgroup
}
    
\begin{document}

\def\sectionautorefname{Sec.}
\def\subsectionautorefname{Sec.}
\def\figureautorefname{Fig.}
\def\tableautorefname{Tab.}
\def\algorithmautorefname{Alg.}
\def\equationautorefname{Eq.}

\title{TabVFL: Improving Latent Representation in Vertical Federated Learning
}


\author{

   \IEEEauthorblockN{Mohamed Rashad\IEEEauthorrefmark{1}\thanks{\IEEEauthorrefmark{2} Contact: z.zhao@nus.edu.sg},  
   Zilong Zhao\IEEEauthorrefmark{2}, 
   J\'er\'emie Decouchant\IEEEauthorrefmark{1}, Lydia Y. Chen
    \IEEEauthorrefmark{1}}
    \IEEEauthorblockA{\IEEEauthorrefmark{1}Delft University of Technology, The Netherlands} 
    \IEEEauthorblockA{\IEEEauthorrefmark{2}National University of Singapore, Singapore} 
}

\maketitle

\thispagestyle{plain}

\begin{abstract}
\blfootnote{This document is a preprint of a paper accepted at IEEE SRDS 2024.}
Autoencoders are popular neural networks that are able to compress high dimensional data to extract relevant latent information. TabNet is a state-of-the-art neural network model designed for tabular data that utilizes an autoencoder architecture for training. 
Vertical Federated Learning (VFL) is an emerging distributed machine learning paradigm that allows multiple parties to train a model collaboratively on vertically partitioned data while maintaining data privacy.
The existing design of training autoencoders in VFL is to train a separate autoencoder in each participant and aggregate the latent representation later. This design could potentially break important correlations between feature data of participating parties, as each autoencoder is trained on locally available features while disregarding the features of others.
In addition, traditional autoencoders are not specifically designed for tabular data, which is ubiquitous in VFL settings. 
Moreover, the impact of client failures during training on the model robustness is under-researched in the VFL scene.
In this paper, we propose \algo, a distributed framework designed to improve latent representation learning 
using the joint features of participants.
The framework (i) preserves privacy by mitigating potential data leakage with the addition of a fully-connected layer, (ii) conserves feature correlations by learning one latent representation vector, and (iii) provides enhanced robustness against client failures during training phase. Extensive experiments on five classification datasets show that \algo can outperform the prior work design, with 26.12\% of improvement on f1-score. 


\end{abstract}


\section{Introduction}

\begin{figure*}[ht]
     \centering
     \begin{subfigure}[b]{0.4\linewidth}
            \centering
            \includegraphics[width=1\textwidth]{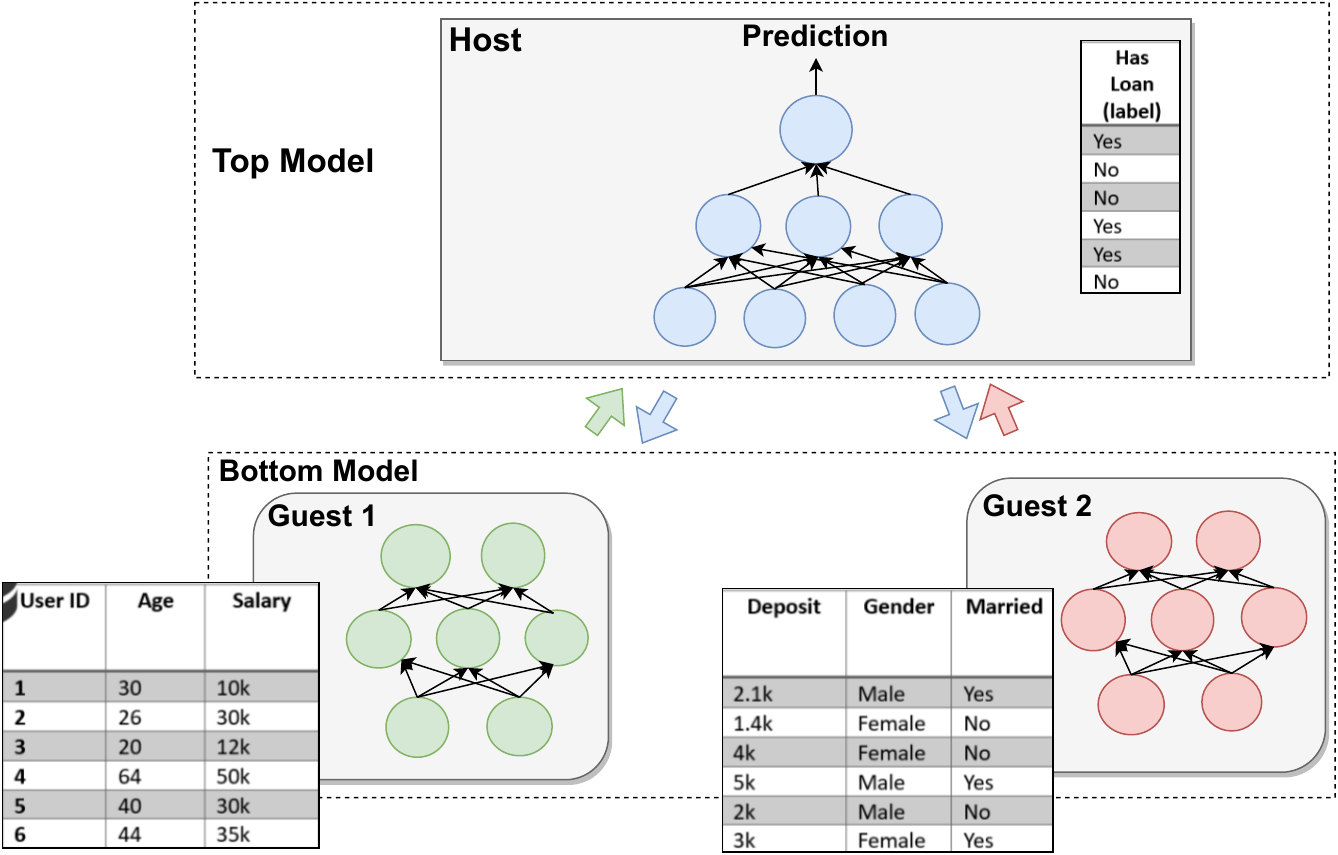}
            \caption{Classical SplitNN Architecture.}
            \label{fig:splitnn}
     \end{subfigure}
     \hfill
     \begin{subfigure}[b]{0.51\linewidth}
            \centering
            \includegraphics[width=1\textwidth]{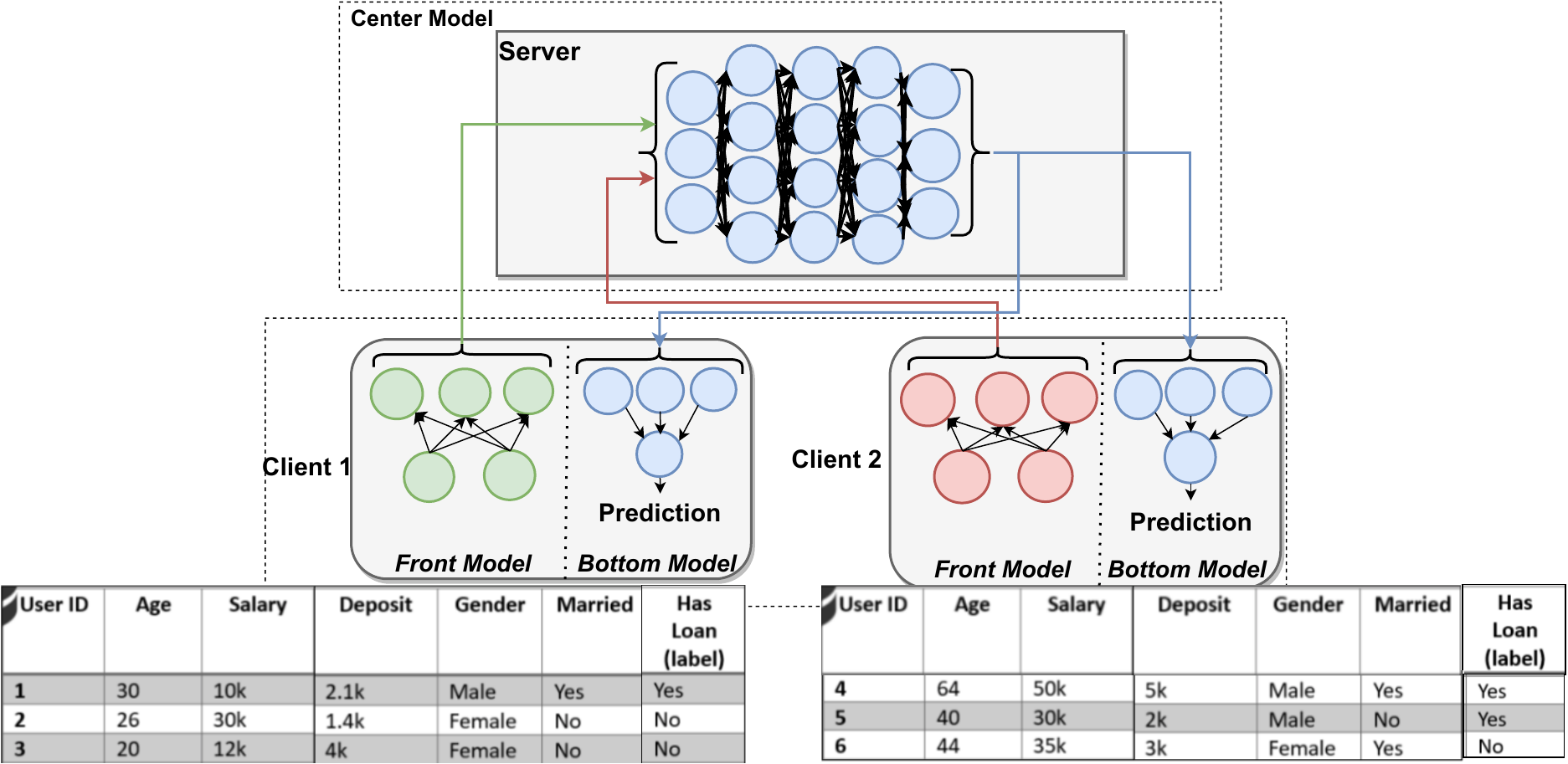}
            \caption{SplitNN U-Shape Architecture.}
            \label{fig:splitnn_ushape}
     \end{subfigure}
     \hfill
        \caption{Predictive model designs using SplitNN: example with two clients.}
         \vspace{-1em}
        \label{fig:splitNN_examples}
\end{figure*}

Autoencoder \cite{Rumelhart1986ParallelDP} is a type of artificial neural network used for unsupervised learning of efficient codings. The main goal of an autoencoder is to learn a representation (encoding) for a set of data, typically for the purpose of dimensionality reduction or feature extraction.  However, despite their success, traditional autoencoders are not specifically designed to handle tabular data effectively. Deep Neural Networks (DNNs) have achieved remarkable results in various domains, especially where data has inherent hierarchical or spatial structures, such as images, audio, and text~\cite{Devlin2019BERTPO,Hinton2012DeepNN,Krizhevsky2012ImageNetCW}. However, when it comes to tabular data, DNNs often underperform compared to boosting-based methods like Gradient Boosted Trees (GBT) \cite{Chen2016XGBoostAS,Ke2017LightGBMAH,Ostroumova2017CatBoostUB}. Tabular data often lacks the spatial or temporal hierarchies that DNNs particularly adapt at modeling. Additionally, DNNs can be more data-hungry and prone to overfitting, especially when the dataset is small, whereas boosting methods can be more robust and generalize better with fewer data points.

Innovative methods are needed to appropriately adapt DNNs to tabular data in order to address these issues. The incorporation of TabNet \cite{Arik2019TabNetAI}, a specialized neural network architecture created specifically for tabular data, is a noteworthy development in this domain. With the use of TabNet, tabular datasets can be handled effectively while utilizing the strength of deep learning. Its unique attention mechanism allows it to focus on relevant subsets of features, enabling the model to capture intricate patterns even in the absence of spatial or temporal properties of the data. 

Vertical Federated Learning (VFL) \cite{Yang2019ParallelDL} describes situations where multiple clients have different features for the same set of individuals. This federated learning approach is especially beneficial when diverse stakeholders or institutions possess distinct data attributes. Consider a bank and an e-commerce platform: while the bank maintains comprehensive financial records, the e-commerce platform holds data on customers' online shopping activities. Both parties aim to collaboratively develop a precise credit scoring model to gauge the risk of a customer defaulting on a loan or credit payment, without directly sharing customer data. In such scenarios, VFL becomes an invaluable tool.


Autoencoders in VFL are substantially under-researched. In the current designs~\cite{xgboost_autoencoder_vfl,Cha2021ImplementingVF,Chu2021PrivacyPreservingSF,Khan2022CommunicationEfficientVF}, each non-label holder is assumed to have a local autoencoder model which is trained only on local feature data. After local training, the learnt latent representation is sent to the label owner. Once received, the latent representations are concatenated and used for downstream tasks, e.g., for training a prediction model. The correlation between the isolated features among the non-label holders is not captured due to local latent representation learning of separate autoencoders. This lack of correlation could lead to the generation of irrelevant or noisy latent data, significantly impacting the prediction performance. To fill in this gap, we propose a joint latent representation learning design, \algo, that learns one latent representation from capturing feature dependencies across all parties involved in a privacy-preserving fashion. SplitNN ~\cite{Poirot2019SplitLF,Vepakomma2018SplitLF} approach is leveraged to assimilate TabNet in the VFL context. This entails the devision of TabNet  into independent model parts that are assigned to each participating party.

Splitting TabNet is not a trivial task due to various dependencies between components. Careful design choices needed to be made in order to protect against the exposure of raw feature data. 
A fully-connected layer is added in each non-label owner model part to prevent the potential occurrence of a direct data leakage.

One major challenge that any federated learning system can suffer from is the presence of failing clients ~\cite{Ang2019RobustFL,Chen2022CFLCF,Ceballos2020SplitNNdrivenVP}. These failures may happen for a number of reasons, including network problems, hardware problems, or software crashes. Client failures can disrupt the training and hence the performance of the system. A naive solution to this problem is replacing missing values of failed clients with vectors of zeros \cite{Sun2023RobustAI}. However, this adds bias to the training and can hinder the performance. We instead propose a caching mechanism for storing and reusing the communicated values to decrease performance degradation and enhance training stability.

We conduct extensive evaluation on \algo using five classification datasets. The results are compared with three baselines.
The accuracy, f1-score and ROC-AUC metrics are reported. The evaluation results show that \algo performs considerably better than prior work for all the datasets. The highest improvement reached was 26.12\% on f1-score. 
\algo outperforms other designs in terms of runtime and memory utilization but it has higher communication overhead compared to the baselines.          
The main contribution of this study can be summarized as follows:
\begin{itemize}[leftmargin=*]
    \item We design the \algo distributed framework, which 
incorporates the state-of-the-art tabular model TabNet into vertical federated learning (VFL) via a single latent representation of vertically partitioned data.
    \item We incorporate an additional fully-connected layer into \algo to prevent potential data leakage. 
    \item We introduce a caching mechanism to mitigate the impact of client failures, enhance training process stability and minimize performance degradation
    \item Extensive experiments are conducted on five classification datasets. Results show that \algo demonstrates significant enhancement in latent quality, outperforming previous designs. 
    Additionally, \algo excels in runtime and memory efficiency, and only incurs a moderately increased network consumption compared to other designs.
\end{itemize}



\section{Preliminaries} \label{sec:prelim}

\textbf{Vertical Federated Learning (VFL)} \cite{Yang2019ParallelDL} is a distributed machine learning paradigm that focuses on collaborative model training between multiple parties that provide their locally available data while maintaining data privacy. The parties are distinguished into two types: guest clients and a host client. The host client acts as the server or the federator and the guests act as feature providers. It is generally assumed that the host is the label holder. The features and the corresponding target labels of the parties need to be aligned prior to the training process. Private Set Intersection (PSI) (\cite{Chen2017FastPS}, \cite{Lu2020MultiPartyPS}) is the usual algorithm applied for the cryptographic alignment of samples in VFL (\cite{Angelou2020AsymmetricPS}, \cite{Mao2023FullDP}). 
We assume that the samples among clients are already aligned and that each client owns distinct features. 
Popular machine learning models for tabular data learning, such as gradient boosting decision trees (GBDTs) (\cite{Wu2020PrivacyPV,Cheng2019SecureBoostAL}) and generative models (\cite{zhao2023gtv}), have been considered in VFL. Participants in tree-based models collaborate to collectively construct a prediction tree, while generative models like GANs are employed for synthesizing tabular data that mimics the patterns in participants' original data.

\begin{figure}[t]
    \centering
    \includegraphics[width=0.5\textwidth]{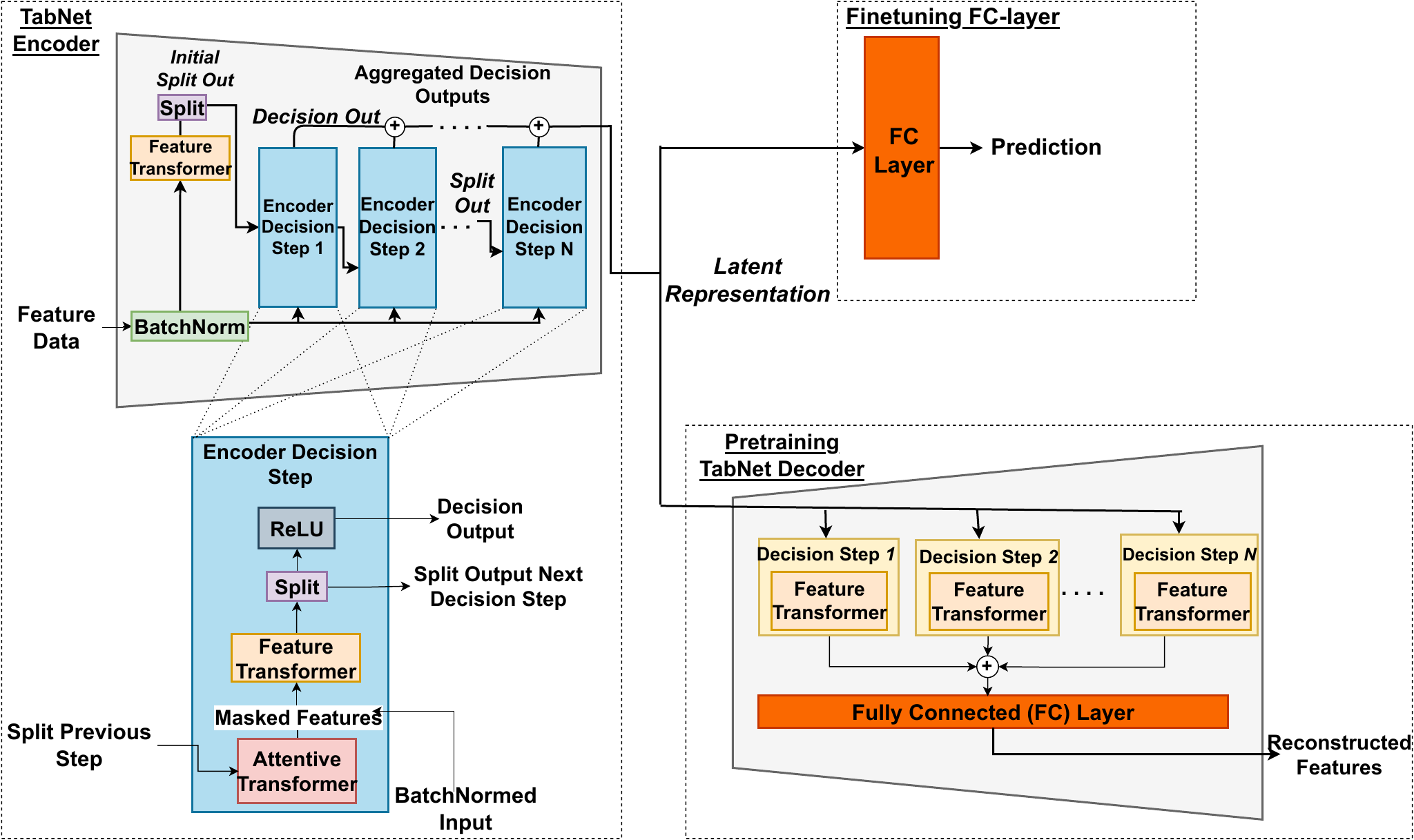}
\caption{Structure of TabNet.}
     \vspace{-1em}
    \label{fig:tabnet_model}
\end{figure}

\textbf{SplitNN} particularly focuses on the neural network based predictors~
\cite{Poirot2019SplitLF,Vepakomma2018SplitLF}. SplitNN enables different and flexible split configurations of a model.
Typically, a splitNN model consists of a bottom and a top model as shown in \autoref{fig:splitnn}. Each guest client processes the local feature data through its bottom model. The generated intermediate results are sent by the guests to the host client. Aggregation takes place of the intermediate results which is further processed by the top model at the host. Finally, a prediction is made and compared to the true value for loss calculation. The acquired loss value is used to update the models by calculating the appropriate gradients.
A special kind of split learning design is the U-shaped design shown in \autoref{fig:splitnn_ushape}. 
The parties are called server and client since they don't adhere to the assumed guest and host roles.
Clients posses a front and bottom model parts, while the server holds the middle part. This naturally reduces computational cost for clients. The front model creates initial intermediate results from input data. Results from clients are sent to the server for processing. Server-generated results are then sent back to clients for prediction and loss calculation.
\textbf{TabNet} \cite{Arik2019TabNetAI} shown in \autoref{fig:tabnet_model} is a state-of-the-art deep neural network that is specifically designed for tabular data. It combines advanced ensembling concepts and novel sequential attention mechanisms. TabNet training phases consists of pretraining and finetuning. During pretraining, an autoencoder \cite{Rumelhart1986ParallelDP} 
structure is employed 
to learn general latent feature representations. The encoder compresses input features, while the decoder reconstructs them, extracting valuable latent features.
For finetuning, the decoder is disregarded and a fully-connected (FC) layer is placed directly after the encoder for predictions. The goal is to specialize the latent representation on a downstream task.
TabNet dynamically learns to pick important features through the attentive layer by creating and applying feature masks 
in each encoder decision step, filtering irrelevant features. 
The encoder outputs sequential decisions that are processed by the decoder for predicting feature values. During finetuning, the encoder decision outputs are transformed into encoded representations instead.

\section{Related Work}
In this section, we begin by introducing tree-based prediction models tailored for tabular data within the VFL training framework. Subsequently, we delve into the exploration of deep learning frameworks highlighted in the VFL literature. Finally, client failure handling methods in federated learning frameworks are discussed.


\subsection{Tree-based Prediction Frameworks In VFL}
Pivot \cite{Wu2020PrivacyPV} introduced an enhanced communication protocol to mitigate data leakage from intermediate results for tree models. SecureBoost \cite{Cheng2019SecureBoostAL} is able to train a GBDT model, such as XGBoost, without sacrificing model utility by performing similar to non-federated settings. The mentioned frameworks are not efficient due to the sequential nature of training the trees. VF$^{2}$Boost \cite{Fu2021VF2BoostVF} proposed a scheduler-worker design to enable concurrent computations and an ad-hoc cryptography protocol for faster communication time. A random forest \cite{Liu2019FederatedF} has also been considered in VFL using a trusted third-party to build and achieve a lossless performing model. 

TabNet~\cite{Arik2019TabNetAI} shows superiority over tree-based model on tabular data predictions. We harness TabNet's predictive prowess, integrating it with splitNN to undertake prediction tasks within the VFL training framework.

\subsection{Deep Learning Prediction Frameworks In VFL}

SplitNN
is applied in multiple VFL use cases, such as healthcare~\cite{Poirot2019SplitLF, Vepakomma2018SplitLF, Roth2022SplitUNetPD}, advertisement \cite{Li2022VerticalSL} and finance \cite{zhao2023gtv}. In \cite{Li2022VerticalSL}, transfer learning is leveraged on overlapped and non-overlapped data among participants to alleviate the issue of missing features. Different configurations for split learning are proposed in \cite{Vepakomma2018SplitLF,zhao2023gtv} depending on different use cases and requirements. Different aggregation methods of intermediate results have been explored specifically in 
VFL context \cite{Ceballos2020SplitNNdrivenVP}. Practical application of splitNN is employed in the PyVertical framework \cite{Romanini2021PyVerticalAV} where a hybrid approach is utilized to internalize split learning in VFL along with the PSI method for entity alignment. FDML \cite{Hu2019FDMLAC} is another VFL framework that also supports neural networks but assumes label availability for all parties. 





FedOnce framework \cite{Wu2022PracticalVF} enforces one communication round between the guests and host clients using a splitNN design. This is achieved by locally pretraining guest models using a technique called \textit{Noise As Targets} to generate latent representations once.
Khan et al. \cite{Khan2022CommunicationEfficientVF} follows the same procedure but a local autoencoder is assumed in each guest client for learning a compressed feature representation.
After completion, the latent data is aggregated and prepared for training a separate prediction model in the host side, e.g., using XGBoost \cite{xgboost_autoencoder_vfl} on the extracted features. A similar autoencoder design is employed in \cite{Cha2021ImplementingVF} with the only difference being that the autoencoder is encouraged to learn a higher dimensional representation of the data.
In \cite{Chu2021PrivacyPreservingSF} a variational autoencoder is used for learning a latent representation that closely represents a Gaussian distribution in order to ensure statistical similarity between the latent data of different samples, hindering sample distinction by attackers. 

Our proposed framework extends the current autoencoder design in VFL by unifying the latent data through training one autoencoder model during pretraining of TabNet. A U-shaped splitNN design (as shown in \autoref{fig:splitnn_ushape}) is utilized for joint learning of the feature representations across parties. To the best of our knowledge, none of the existing VFL frameworks considers the application of U-shaped splitNN in cases of feature reconstruction.


\subsection{Client Failure Handling Methods}
Many client failure handling methods exist in the Horizontal Federated Learning (HFL) context where it is assumed that clients have overlapping feature space but distinct data samples. Existing techniques include  alleviation of client failures by incorporating robust aggregation algorithms \cite{Ang2019RobustFL}, \cite{Chen2022CFLCF}, increasing eligibility of participating devices in scalable environments \cite{Bonawitz2019TowardsFL}, preemptive client state prediction \cite{Yang2021CharacterizingIO} and by dynamically varying training round duration \cite{Li2019SmartPCHP}. An empirical study has also shown that client failures have significant impact on convergence time and performance of an FL system \cite{Yang2020HeterogeneityAwareFL}. However, literature about this matter in VFL is scarce~\cite{Ceballos2020SplitNNdrivenVP,Sun2023RobustAI,Hou2021VerifiablePS}. Iker Ceballos et al. \cite{Ceballos2020SplitNNdrivenVP} demonstrated that quitting of clients leads to slower convergence rates and worse performance, amplified by simultaneous quitting of clients in a four-client splitNN configuration. Client failures in a two-party splitNN setting are also conducted where a dropout-based method is introduced to enforce robustness against a quitting client \cite{Sun2023RobustAI}.
Furthermore, \cite{Hou2021VerifiablePS} achieves model robustness via dynamic client data re-indexing. In this study, we leverage a caching method to improve model robustness in cases of client failures within a two-party setup.

\section{Methodology} 
In this section, we first delve into the architecture of \algo, elucidating its design and rationale. We then address crucial privacy aspects of the model. Lastly, we present the training specifics, highlighting a novel approach to manage client failures.

\begin{table}[t]
    \centering
    \caption{Notations}
    \label{tab:symbols}
    \begin{tabular}{c p{2in}}
        \hline
        \textbf{Symbol} & \textbf{Description} \\
        \hline
        \textit{PartialEnc} & The encoder module of TabNet without BN layer residing in the host client.\\
        
        \textit{PartialDec} & The decoder module of TabNet without the FC layer residing in the host client.\\
        
        \textbf{$f_{\mathrm{bottom\_repr}}$} & The feature extractor containing a BN and a FC layer in the bottom model of guest clients.\\
        
        \textbf{$f_{\mathrm{bottom\_repr\_bin}}$} & The feature extractor containing a BN, FC and RandomObfuscator layers in the bottom model of guest clients.\\
        
        \textbf{$f_{\mathrm{bottom\_rec}}$} & FC layer used for feature reconstruction in the bottom model of guest clients.\\
        
        \textbf{$final\_mapping$} & The final FC layer in the finetuning model of \algo used for making a prediction in the host client.\\
        \hline
    \end{tabular}
    \vspace{-0.5em}
\end{table}


\begin{figure*}[h!]
     \centering
     \begin{subfigure}[b]{0.5\linewidth}
            \centering
            \includegraphics[width=1\linewidth]{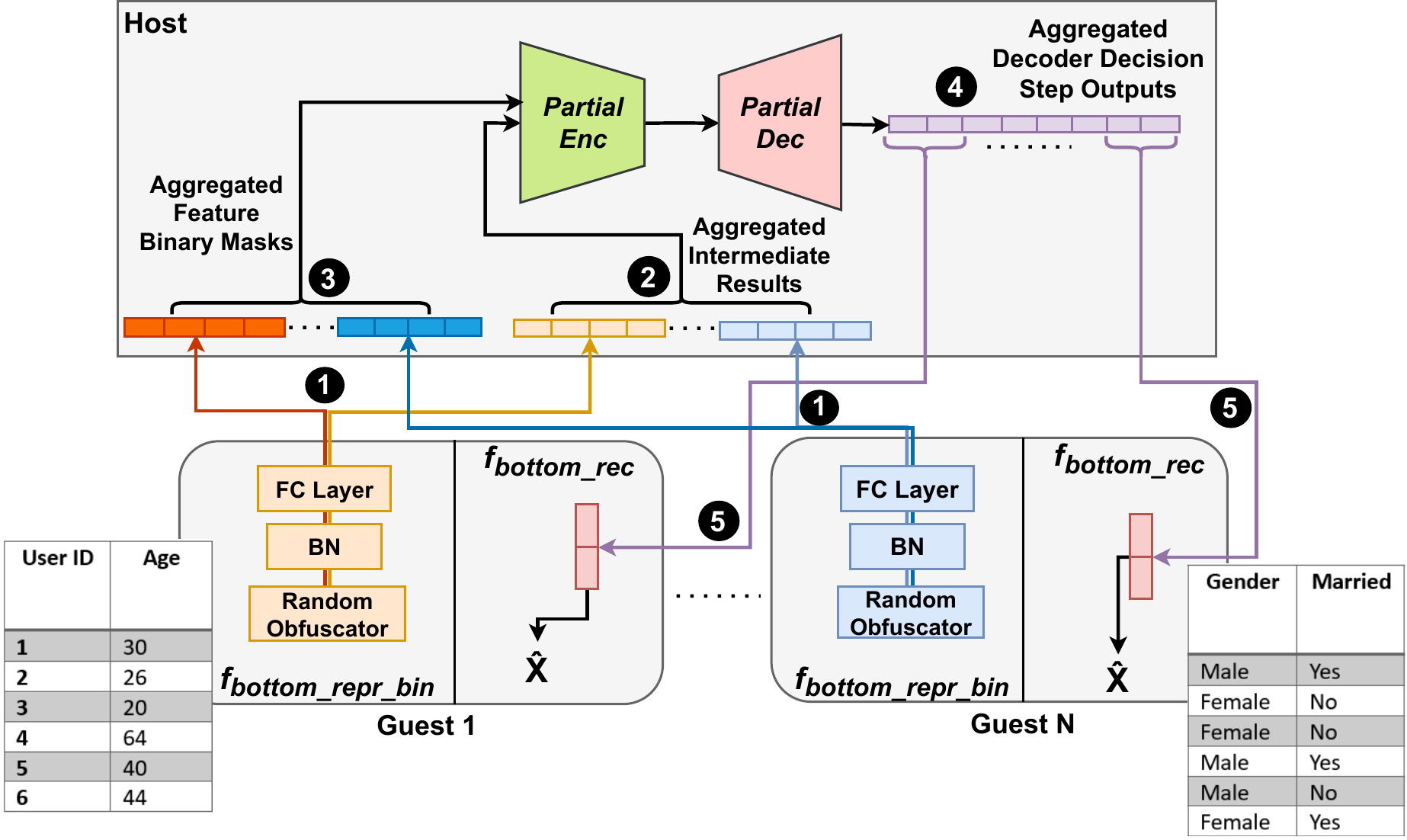}
            \caption{\algo pretraining workflow.}
            \label{fig:tabvfl_pretraining}
     \end{subfigure}
     \begin{subfigure}[b]{0.42\linewidth}
            \centering
            \includegraphics[width=1\linewidth]{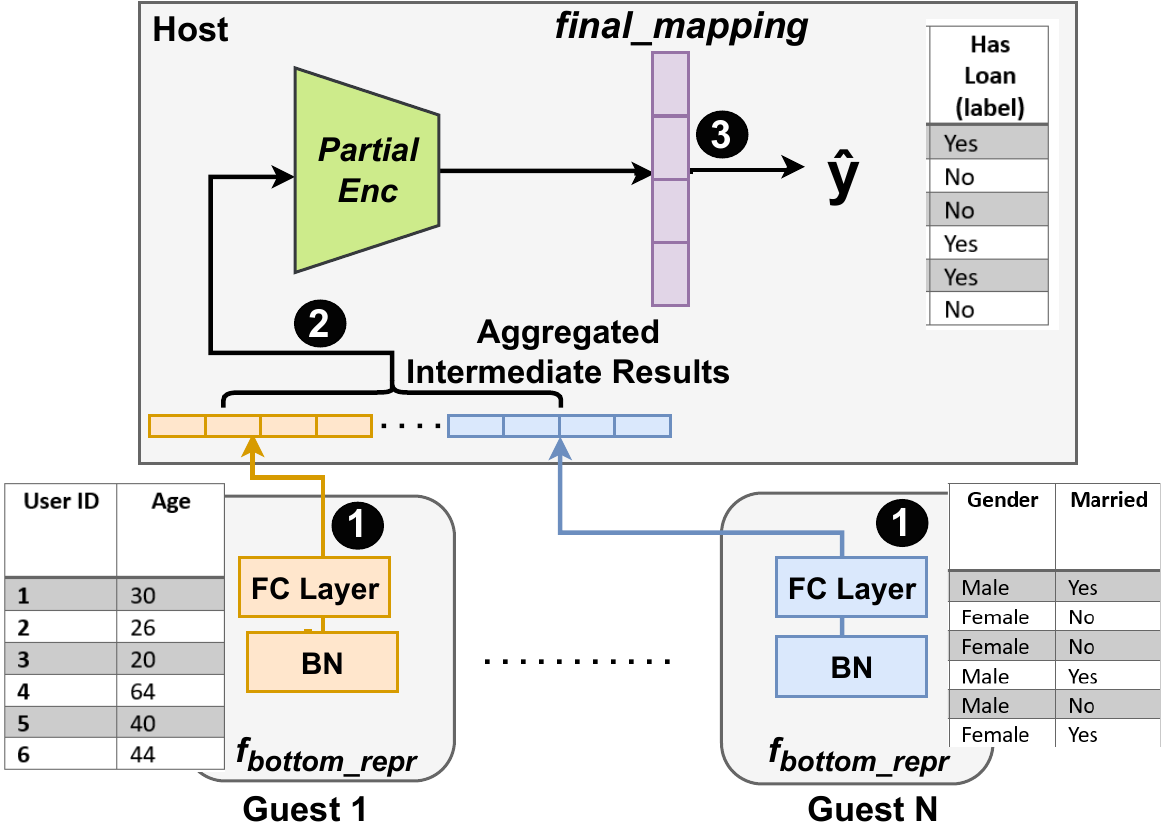}
            \caption{\algo finetuning workflow. }
            \label{fig:tabvfl_finetuning}
     \end{subfigure}
     \hfill
        \caption{Overview of pretraining and finetuning workflow in \algo.}
        \vspace{-1em}
        \label{fig:combined_tabnet_training_stages}
\end{figure*}

\subsection{TabVFL Architecture Overview}
In \autoref{fig:combined_tabnet_training_stages}, the structure of \algo is shown in a two client setup with an example dataset. Each guest client owns distinct feature data corresponding to the same samples. The novel design is introduced in the pretraining phase of \algo. The encoder and decoder components of the TabNet model are split into two parts: the top and bottom models. The majority of the model is retained in the host client. This is done to be consistent with the general model design of U-shaped neural network shown in \autoref{fig:splitnn_ushape}. 

Many components of the TabNet pretraining model are dependent due to the sequential attention structure which makes a proper splitting of the model among the parties difficult (\autoref{fig:tabnet_model}). Nonetheless, the encoder can be conveniently split by assigning the first batch normalization (BN) layer of the TabNet encoder to the guest clients and the remaining encoder components to the host client. The reason can be deduced from the valuable observation that the BN layer is the only part that every other component depends on since the normalized values are passed to each encoder decision step and to the initial split. Moreover, the decoder can be sufficiently divided by splitting it between the feature transformers and the fully-connected (FC) layer. The FC-layer is further split uniformly into partitions such that each partition corresponds to each guest client. The FC-layers are held locally in each guest. The host only instructs the guests to initialize their FC-layer with the partition dimension. The output dimension of the feature transformers blocks is equal to the latent dimension. Therefore, it should be ensured that the latent dimension is not set to a value that is lower than the number of participating guests. The top model consists of \textit{PartialEnc} and \textit{PartialDec}. The \algo pretraining model is shown in \autoref{fig:tabvfl_pretraining}.

The bottom model is represented by two parts: the feature extractor (left part of the bottom model) and the feature reconstructor (right part of the bottom model). The feature extractor is used to transform the raw feature data into intermediate logits. The \textit{RandomObfuscator} is a non-learnable component used to generate binary masks which are directly applied to the features. The reason for the additional FC-layer is to transform the outputs of the BN layer in case it behaves as an identity function. This is possible when the features are already standardized and follow a standard normal distribution, rendering the BN layer useless for protecting against direct data leakage.
The feature reconstructor is there to predict the actual feature values that were used as input. The split of the decoder is crucial for preventing the host client from reverse engineering the features of the guest clients. This is possible if the host client has a particular amount of feature samples of some or all the guest clients. The encoder split was done to allow general latent representation learning of one latent vector by aggregating the intermediate logits of all guest clients at the host. The concept of a label is absent during pretraining since the goal is to accurately reconstruct the masked features resulted from the \textit{RandomObfuscator}.   

In the finetuning design, the host is the label owner. The conventional splitNN procedure is followed in order to specialize the latent representation from the encoder to a specific task. As shown in \autoref{fig:tabvfl_finetuning}, the decoder components  are disregarded including the \textit{RandomObfuscator} component as they are not needed for finetuning. 


\subsection{Threat Model}

We assume honest-but-curious setting where all participants follow the framework protocol properly without any (malicious) deviation. Specifically, we assume that the guest and host clients are honest but curious, meaning that they adhere to the \algo protocol, but may seek to acquire additional information through the computation process ~\cite{Fang2020LargescaleSX,Fu2022BlindFLVF}. We do not consider colluding host and guests. Since the model weights are initialized locally and only intermediate results are exchanged, the models in each client are considered as black box to other clients.

\subsection{TabVFL Training Procedures}
Identical training steps as in the TabNet model are followed in \algo. In both training phases, the host client is responsible for federating the training process in mini-batch fashion. The training workflow for pretraining and finetuning are provided with an example dataset in \autoref{fig:combined_tabnet_training_stages}. Similar step numbers indicate parallel computations/communications. 

During the pretraining phase, the first step entails the generation of the binary masks and the transformed representation of the raw data by the guest clients (\stepnum{1}). The batch of raw features is first masked using the generated masks from \textit{RandomObfuscator} which are processed by a BN and a FC layer and transmitted to the host. The binary masks and intermediate results from the guest clients are aggregated by concatenating them (\stepnum{2}, \stepnum{3}). The concatenated values are processed by the \textit{PartialEnc} and passed to the \textit{PartialDec} to generate intermediate decoder representations by aggregating feature transformers outputs, which are then partitioned (\stepnum{4}). Each partition result is sent to its corresponding guest client for reconstructing the masked feature values (\stepnum{5}). The reconstruction values are used to calculate the gradients for updating \textit{$f_{\mathrm{bottom\_rec}}$}, \textit{PartialDec}, \textit{PartialEnc} and \textit{$f_{\mathrm{bottom\_repr\_bin}}$}.

During finetuning, the raw features are processed by each guest client through the pretrained BN and FC layers and communicated to the host (\stepnum{1}). Subsequently, the intermediate results are gathered and concatenated (\stepnum{2}). The concatenated results are fed through the \textit{PartialEnc} to generate the latent representations which are processed by the \textit{$final\_mapping$} for predicting the corresponding label for each sample (\stepnum{3}). The predicted value is used to calculate the gradients for updating \textit{$final\_mapping$}, \textit{PartialEnc} and \textit{$f_{\mathrm{bottom\_repr}}$}.

In each training phase, one epoch/iteration is finished once the weights are updated by executing the described workflow.

\subsection{Training Details}
The details of the training phases for pretraining and finetuning are presented in \autoref{alg:tabnet_pretrain_algo} and \autoref{alg:tabnet_finetune_algo}. The training of \algo starts off with the pretraining phase. Each guest first initializes its own bottom models \textit{$f_{\mathrm{bottom\_repr\_bin}}$} and \textit{$f_{\mathrm{bottom\_rec}}$} (line \textbf{1}). Simultaneously, the host initializes the \textit{PartialEnc} and \textit{PartialDec} models (line \textbf{7}). The guest clients start the forward propagation by generating binary masks and apply them to the raw feature data. The masked batch is processed by \textit{$f_{\mathrm{bottom\_repr\_bin}}$} model and sent to host (lines \textbf{4}-\textbf{6}). The host receives the intermediate results and the binary masks which are both concatenated into one vector separately (lines \textbf{13}-\textbf{16}). The results are also cached for the client handling method described in \autoref{subsec:caching_method}. Although the host can generate random binary masks itself, the binary masks are cached instead as well to not break its correlation with the corresponding intermediate results. The concatenated values are fed to the \textit{PartialEnc} to generate the corresponding latent vector. The latent vector is in turn processed by the \textit{PartialDec} to create the intermediate decoder representations (lines \textbf{17}-\textbf{18}). The decoder results are split into uniform chunks where each chunk corresponds to one guest client. The chunks are sent to each guest for reconstruction (lines \textbf{19}-\textbf{21}). The guests receive their intermediate decoder results and reconstruct the masked features using \textit{$f_{\mathrm{bottom\_rec}}$}. The output is compared with the actual feature values for calculating the reconstruction loss (lines \textbf{22}-\textbf{24}). For backpropagation, one loss value is required. To achieve this, each guest client is ordered to send its loss to the host. The accumulation of the losses takes place and the result is used to calculate the gradients and update the weights of the top and bottom models (lines \textbf{25}-\textbf{32}). A modified version of the MSE function is used in TabNet to account for the binary mask and feature ranges for a more stable loss calculation.


During finetuning, the weights of the FC and BN layers in \textit{$f_{\mathrm{bottom\_repr\_bin}}$} are copied over in \textit{$f_{\mathrm{bottom\_repr}}$} for further finetuning (line \textbf{1}). The host now uses the pretrained \textit{PartialEnc} and initializes a \textit{final\_mapping} containing an FC layer for prediction (line \textbf{5}-\textbf{6}). The first step of the forward pass is sending the intermediate results, from applying \textit{$f_{\mathrm{bottom\_repr}}$} to the raw features, to the host (line \textbf{4}). The host receives the transmitted data from each guest client and caches it to handle possible client failures (line \textbf{11}). Similar to pretraining, a vector with concatenated intermediate results is created which is passed to the \textit{PartialEnc}. The \textit{PartialEnc} now also generates \textit{M\_Loss} along with the latent vector (lines \textbf{12}-\textbf{13}). The final logits are created from passing the latent vector through the \textit{final\_mapping} layer. Afterwards, the final logits are softmaxed to create normalized prediction probabilities from logits. The loss is calculated by comparing the prediction probabilities and the actual label values (lines \textbf{14}-\textbf{16}). Finally, the \textit{M\_loss} is multiplied by the regularization parameter  \textit{$\lambda_{sparse}$}. The \textit{M\_loss} is used for regularizing the sparsity of the feature selection made by the attentive layer of TabNet. This is to encourage the model to only pick important features for datasets that have few features that are relevant.
The final loss value is used to calculate gradients and update the weights accordingly (lines \textbf{17}-\textbf{18}). The classification loss represents the cross entropy. 

Note that the host is considered to be the federator which instructs the guest clients to send and receive data during training and inference.

\begin{algorithm}
\DontPrintSemicolon
\SetCommentSty{commfont}
\KwIn{  

    NumClients = $\{1, 2, ... K\}$  
    
    Epochs = $E$  
    
    Feature Dataset $X_{i}$ \text{of party} $i$ $\in$ $\{2, 3, ... K\}$  
    
}
\tcc{\textbf{Guests:}}
    \textbf{initialize} local models $f_{\mathrm{bottom\_repr\_bin}}, f_{\mathrm{bottom\_rec}}$

    \For{$e \in E$} {
        \For{each mini-batch $X_{b} \in B$} {
            
            Generate binary mask $S_{b}$

            $X_{b}^{masked}$ $\leftarrow$ Mask $X_{b}$ using $S_{b}$ 

            Send intermediate result $f_{\mathrm{bottom\_repr\_bin}}(X_{b}^{masked})$ and $S_{b}$ to host

\tcc{\textbf{Client 1 (Host)}}
        \textbf{initialize} $PartialEnc, PartialDec$
    
        \For{$c \in [2,K]$} {
            \If{c is offline}{
                $\Tilde{X} \leftarrow $ reuse cached intermediate results 
                
                $\Tilde{S} \leftarrow $ reuse cached binary masks 
            }
            \Else{
                $\Tilde{X} \leftarrow$ Receive and append intermediate results. Cache received results. 
            
                $\Tilde{S} \leftarrow$ Receive and append partial binary masks. Cache received binary masks. 
            }
        }
        
        $X_{intermediate} \leftarrow$ concatenate $\Tilde{X}$ column-wise
    
        $S_{complete} \leftarrow$ concatenate $\Tilde{S}$ column-wise
    
        $Z_{latent} \leftarrow PartialEnc(X_{intermediate}, S_{complete}$)
    
        $out_{intermediate} \leftarrow PartialDec(Z_{latent})$
    
        $[\Tilde{out}_1, \Tilde{out}_2, ..., \Tilde{out}_K] \leftarrow$ Split $out_{intermediate}$ into uniform partitions for client $c \in [2, K]$
    
        \For{$c \in [2,K]$} {
            Send $\Tilde{out}_c$ to client $c$
        }
    
\tcc{\textbf{Guests:}}
        
        $D_{b} \leftarrow$ Receive mini-batch decoder intermediate result $\Tilde{out}_c$
    
        $\hat{X_{b}} \leftarrow$ $f_{\mathrm{bottom\_rec}}(D_{b})$
    
        $loss_{b} \leftarrow$ $reconstruction\_loss(X_{b}, \hat{X_{b}}, S_{b})$
    
        Send $loss_{b}$ to host 

\tcc{\textbf{Client 1 (Host)}}
        $total\_loss$ = 0
    
        \For{$c \in [2,K]$} {
            \If{client c is offline}{
                continue
            }
            
            $loss_c \leftarrow$ Receive mini-batch unsupervised loss
    
            $total\_loss = total\_loss + loss_c$
        }
        
        Calculate gradients and update the weights
    } 
}
\caption{\algo Pretraining Process}
\label{alg:tabnet_pretrain_algo}
\end{algorithm}

\subsection{Client Failure Handling} \label{subsec:caching_method}
Usually in practical cases, variation in the systems of the guest clients can naturally lead to training errors during the training session. This error can occur when a guest client is unable to complete the training due to an unreliable channel or limited data transfer speed, resulting in a failed connection. We propose a caching method for handling abrupt failures of clients to improve robustness and stability during training. The method is implemented in the host client as that is where the intermediate results are handled of each guest client. 
During training, if a client goes offline then its stored/cached batch of results are reused instead (code lines \textbf{10}-\textbf{11} in \autoref{alg:tabnet_pretrain_algo} and \textbf{9} in \autoref{alg:tabnet_finetune_algo}). When a client is online again, the corresponding batches in the cache are replaced by the new intermediate results (code lines \textbf{13}-\textbf{14} in \autoref{alg:tabnet_pretrain_algo} and \textbf{11} in \autoref{alg:tabnet_finetune_algo}). Therefore, we enforce that the clients must be online during the initial epoch.
To simulate the failures using the proposed framework, the guest clients are given the same probability of being considered offline. Let \textit{M} be the number of guest clients. Each guest client $\textbf{\textit{c}} \in M$ follows a binary failure probability distribution:
\begin{equation*}
\textit{\textbf{p}}_{c} \sim Bernoulli(P\_{fail}) \quad \forall c \in M
\end{equation*}
We simulate the setting by sampling a float value \textbf{\textit{x}} from a random variable that follows a uniform distribution for each client, i.e. $\textbf{X} \sim U(0, 1)$. If the sampled value is less than a predefined threshold ($P\_{fail}$), then the client is skipped for one whole epoch iteration. The skipped clients would not be able to generate any new intermediate results. 
In a practical setting, the client failures could be handled at the host by setting a certain timeout to wait a bit for a guest client to send its intermediate results. If a timeout occurs, then it is assumed that the guest client is offline or unresponsive. Furthermore, the correct batch index that is being processed by the host should also be communicated to the guest clients to keep the mini-batches aligned throughout the training process during a real deployment. This should be done to prevent offline clients from processing the wrong batch when they come online again. 

\begin{algorithm}
\DontPrintSemicolon
\SetCommentSty{commfont}
\KwIn{  

    NumClients = $\{1, 2, ... K\}$  
    
    Epochs = $E$  
    
    Feature Dataset $X_{i}$ \text{of party} $i$ $\in$ $\{2, 3, ... K\}$
}
\tcc{\textbf{Guests:}}
    \textbf{set} weights of $f_{\mathrm{bottom\_repr}}$ to the pretrained BN and FC layers $f_{\mathrm{bottom\_repr\_bin}}$

    \For{$e \in E$} {
        \For{each mini-batch $X_{b} \in B$} {
            
            Send intermediate result $f_{\mathrm{bottom\_repr}}(X_{b})$ to host

\tcc{\textbf{Client 1 (Host)}}
    \textbf{use pretrained} $PartialEnc$
    
    \textbf{initialize} $final\_mapping$

    \For{$c \in [2,K]$} {
        \If{c is offline}{
            $\Tilde{X} \leftarrow $ append cached  intermediate results
        }
        \Else{
            $\Tilde{X} \leftarrow$ Receive and append intermediate results. Cache the received results. 
        }
    }
    
    $X_{intermediate} \leftarrow$ concatenate $\Tilde{X}$ column-wise

    $Z_{latent}, M\_Loss \leftarrow PartialEnc(X_{intermediate}$)

    $final\_logits \leftarrow final\_mapping(Z_{latent})$

    $prediction\_proba \leftarrow softmax(final\_logits)$

    $loss \leftarrow classification\_loss(prediction\_proba, y\_true)$

    $loss = loss - M\_loss * \lambda_{sparse}$
    
    Calculate gradients and update the weights 
    } 
}
\caption{\algo Finetuning Process}
\label{alg:tabnet_finetune_algo}
\end{algorithm}

\section{Performance Evaluation}

In this section, we first introduce the experimental setup and the baselines. We then showcase the evaluation pipeline, followed by the presentation and analysis of the results.


\subsection{Experiment Details} \label{sec:exp_details}
\textbf{Setup.} \algo \text{is} compared against three baseline designs: Central TabNet (\codedtext{CT}), Local TabNets (\codedtext{LT}) and TabNet VFL with guests-assigned encoders (\codedtext{TabVFL-LE}). \codedtext{CT} is the baseline where TabNet is run on the full dataset in a non-federated setting. TabNet model implemented with PyTorch \cite{TabNetGithubRepo} is used. The code is modified to disable batch shuffling in each mini-batch iteration and a feature for measuring epoch time is added. The former is implemented to ensure that the experiments differ only in design.
\codedtext{LT} applies the design from prior work ~\cite{xgboost_autoencoder_vfl,Cha2021ImplementingVF,Khan2022CommunicationEfficientVF,Chu2021PrivacyPreservingSF} where multiple autoencoders are trained locally in each guest client. The TabNet pretraining model is used instead of autoencoders to keep it consistent with the other baselines. During finetuning, one fully-connected (FC) layer is trained collaboratively in the host. This is done by ensuring that only latent representations are sent to the host. 
The latent data of each guest are aggregated to make up one latent vector which is processed by the FC-layer in the host to predict a label and calculate the loss. The decision of training one FC-layer at the host (similar to splitNN) is made to  
fairly compare it with other designs as it resembles the training procedure of finetuning in \algo.
\codedtext{TabVFL-LE} shares the pretraining model design of \algo, differing slightly in that the complete encoder is kept within each guest client, unlike TabVFL where it is split into two parts.
The outputs of all decision steps are sent as a list of tensors to the host (see \autoref{fig:tabvfl_pretraining}) where the output of each decision step is aggregated by means of summation with the outputs of the same decision step from different guests. The reason is that
the decoder accepts a list of encoder decision step tensors. Each encoder decision step corresponds to a decoder decision step.
Due to the host receiving all encoder steps from the guests, concatenating each encoder split output and feeding it to the decoder is not feasible. This approach would create incompatible dimensions for each decoder split.
Consequently, every encoder decision step output received by the host is aggregated into a single vector, ensuring compatibility with the dimensions expected by the decoder.
In \autoref{fig:tabnet_model} it is shown that encoder steps are aggregated and passed to the decoder. But in the PyTorch code of TabNet \cite{TabNetGithubRepo}, the aggregation happens at the decoder. The finetuning phase is identical to \algo with the only difference being that the latent vectors sent from each guest client to the host are aggregated using summation instead of concatenation as in \algo. This is done to keep the aggregation method consistent throughout the training phases. This baseline can be seen as a partial transition from \codedtext{LT} design to \algo \text{design}. We find it interesting to see how the transitioning design affects the evaluation results.

\begin{figure}[t]
    \centering
    \includegraphics[width=0.9\columnwidth]{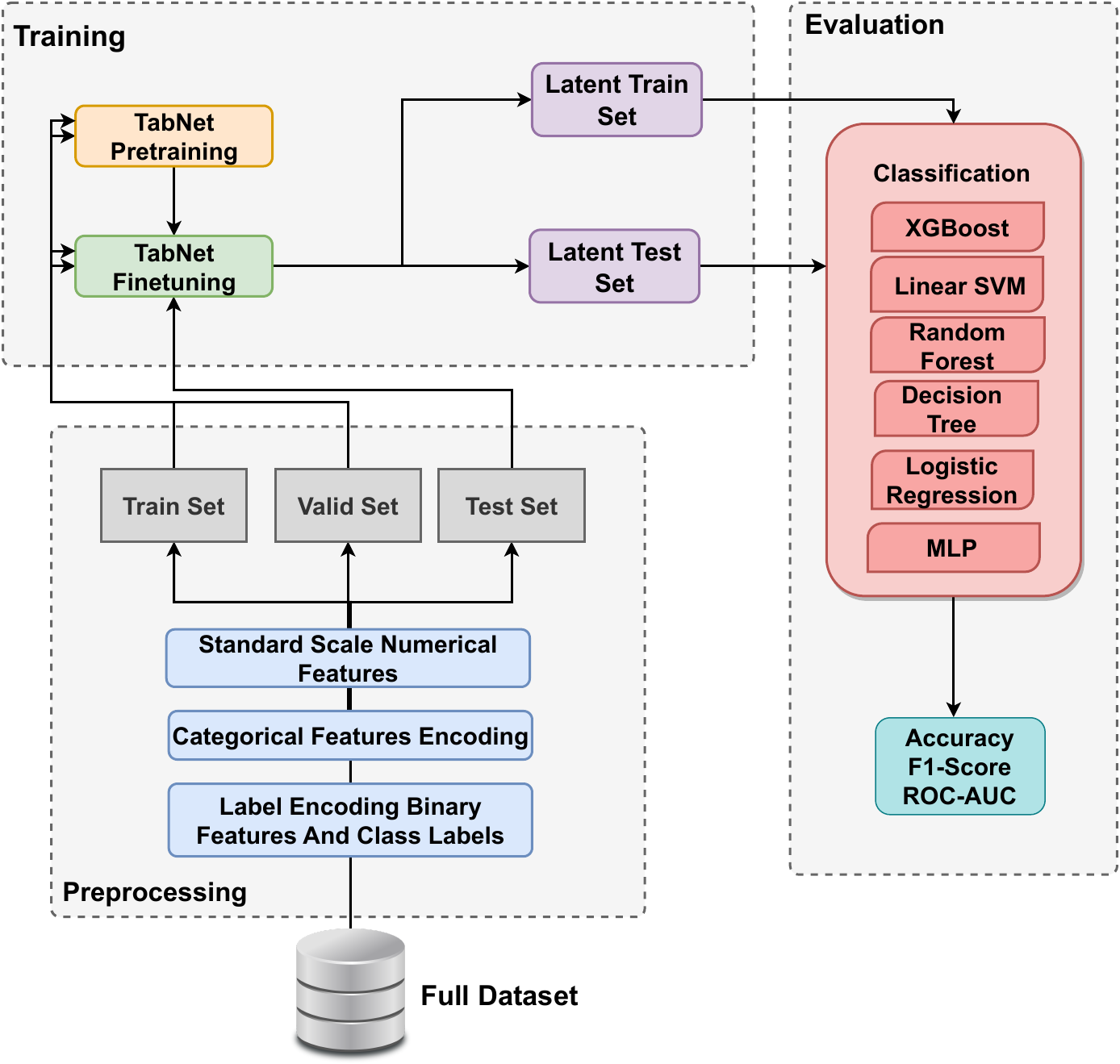}
    \label{fig:evaluation_flowchart}
    \caption{Evaluation pipeline for evaluating the latent data of different experimental designs.}
    \vspace{-1em}
    \label{fig:evaluation_flowchart}
\end{figure}

\begin{figure*}[t]
     \centering
     \begin{subfigure}[b]{0.85\linewidth}
            \centering
            \hspace*{1.5cm}\includegraphics[trim=0cm 0cm 0cm 6.5em, clip, width=1\linewidth]{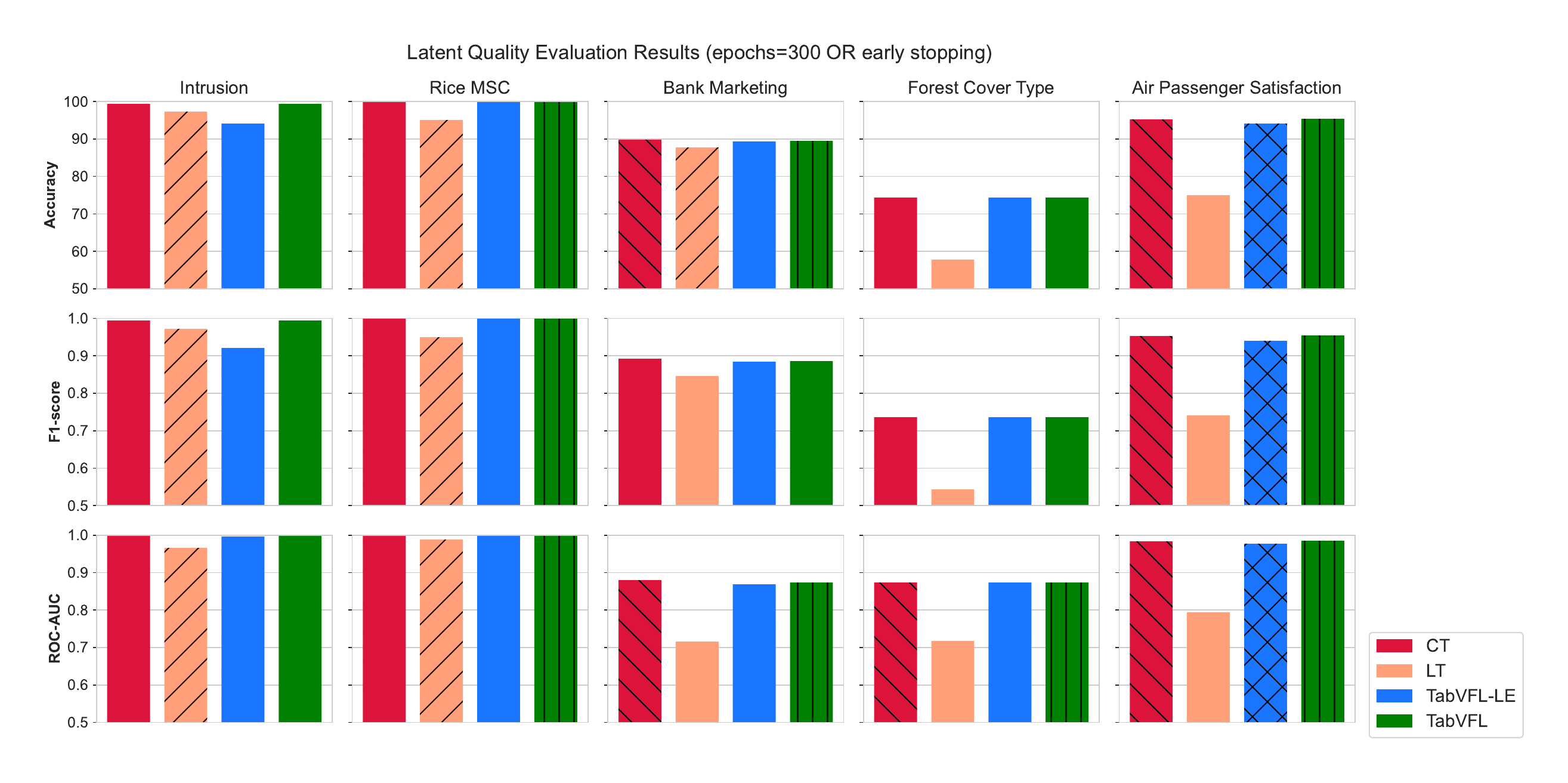}
            \label{fig:latent_eval_result_plot}
     \end{subfigure}
     \hfill
      \vspace{-2em}
        \caption{Results of the latent representation quality of \algo and all baselines on different datasets.}
        \label{fig:latent_eval_results}
\end{figure*}

\textbf{Environment.} The experiments have been run on a machine with NVIDIA GeForce RTX 2080 Ti GPU with 11 GB of VRAM and CUDA version 11.0. The amount of RAM is 32 GB. Intel(R) Core(TM) i9-10900KF CPU is used with speed of 3.70GHz and 20 cores.


\textbf{Datasets.} Five different classification datasets are considered for evaluations: Air Passenger Satisfaction (\codedtext{air}), Network Intrusion Detection (\codedtext{intrusion}), Bank Marketing (\codedtext{bank}), Forest Cover Type (\codedtext{forest}) and Rice MSC (\codedtext{rice}). The datasets  \codedtext{bank} and \codedtext{forest} are retrieved from \href{https://archive.ics.uci.edu/datasets}{UCI Machine Learning Repository}, while \codedtext{air},  \codedtext{intrusion} and \codedtext{rice} are acquired from \href{https://www.kaggle.com/datasets}{Kaggle}. Among the classification datasets, three are suited for binary tasks, while two are tailored for multi-class tasks. 
The metadata of each dataset are shown in \autoref{tab:metadata-datasets-table}. All datasets, with the exception of \codedtext{bank}, are down-sized to 50K samples due to the limited computational resources. Stratified random sampling is employed with regards to the target label to ensure identical class distribution between the original and the down-sized datasets. All the designs are run for 300 epochs to allow ample time for convergence both during pretraining and finetuning. Batch size of 64 is used. 
A latent dimension of five has been chosen for all experiments to facilitate similar model sizes across all baselines and to encourage a compressed latent representation for all datasets. 
{Furthermore, three runs are conducted from which the average is calculated and presented to account for the inherent randomness and variability in the results.} During the experiments, each dataset is split vertically and uniformly to mimic a VFL environment. Moreover, the numerical features of the datasets are preprocessed using standard scaling. Binary features and class labels are label encoded, while the categorical features are one-hot encoded. 


\subsection{Evaluation Pipeline}
An evaluation pipeline has been devised to evaluate the effectiveness of the generated latent representation of each design. The pipeline is shown in \autoref{fig:evaluation_flowchart}. The full dataset is preprocessed and split into training, validation and test sets with a ratio of 70\%-15\%-15\% respectively. The training set is used as input for both pretraining and finetuning in each experiment design. The validation set is used for the early stopping mechanism that we employ during training. A patience value is used as the maximum count of how many iterations to wait for the validation score to improve. The early stopping feature is implemented as a regularization method to prevent overfitting \cite{Caruana2000OverfittingIN}. The training is stopped when the maximum amount of epochs is reached or when the patience value is reached. After training completion, the training, validation and test sets are combined and processed by the TabNet finetuning model to generate the total latent set from the encoder. The set is then split into a training and test set with ratios 70\%-30\%, respectively, using random shuffling to generate the latent train and test sets. The latent train set is used to train six classic machine learning algorithms: logistic regression, decision tree, random forest tree, multi-layer perceptron (MLP), XGBoost and linear support vector machine (Linear SVM).  
The performance, and hence the quality of the generated latent data, of each algorithm is evaluated on the latent test set using the conventional classification metrics: accuracy, F1-score, and the area under the curve of the receiver operating characteristic curve (ROC-AUC). For each experiment run, the mean of the metric results of each algorithm are reported. 


\begin{table}[t]
\tabcolsep=0.2cm
\centering
\caption{Metadata of the datasets used for the experiments.}
\label{tab:metadata-datasets-table}
\resizebox{\columnwidth}{!}{
    \begin{tabular}{|c|c|c|c|c|c|}
    \hline
    Dataset & \#Samples & \#Classes & \#Categorical & \#Numerical & \#Binary  \\ \hline
    Intrusion & 50,000 & 2 & 11 & 23 & 4 \\ \hline
    Rice MSC & 50,000 & 5 & 0 & 106 & 0  \\ 
    \hline 
    Air Passenger Satisfaction & 50,000 & 2 & 15 & 4 & 3  \\ 
    \hline
    Forest Cover Type & 50,000 & 7 & 44 & 10 & 0 \\ \hline
    Bank Marketing & 45,211 & 2 & 6 & 7 & 3\\ \hline
    \end{tabular}
}
\end{table}

\begin{table*}[ht]
  \centering
  \caption{Average evaluation metric results of the six predictors for each latent dimension tested. The results are noted per design. Only F1-scores are included due to lack of space.}
  \label{tab:diff-latent-dims}
      \begin{tabular}{||c|c|c|c|c|c|c|c|c|c|c|c|c||}
        \hline
        \multirow{2}{*}{Dataset} &
        \multicolumn{3}{|c|}{CT} &
        \multicolumn{3}{|c|}{LT} &
        \multicolumn{3}{|c|}{TabVFL-LE} &
        \multicolumn{3}{|c||}{TabVFL}\\
        \cline{2-13}
        & 16 & 32 & 64 & 16 & 32 & 64 & 16 & 32 & 64 & 16 & 32 & 64  \\
        \hline
        Intrusion & 0.996 & 0.993 & 0.996 & 0.919 & 0.982 & 0.989 & 0.85 & 0.992 & 0.947 & 0.995 & 0.993 & 0.995  \\
        \hline
        Rice MSC & 0.998 & 0.998 & 0.999 & 0.994 & 0.996 & 0.997 & 0.998 & 0.998 & 0.999 & 0.998 & 0.998 & 0.999  \\
        \hline
        Bank Marketing & 0.895 & 0.895 & 0.887 & 0.859 & 0.876 & 0.882 & 0.888 & 0.888 & 0.887 & 0.888 & 0.888 & 0.887 \\
        \hline
        Forest Cover Type & 0.772 & 0.792 & 0.78 & 0.636 & 0.635 & 0.7 & 0.76 & 0.771 & 0.78 & 0.772 & 0.785 & 0.78 \\
        \hline
        Air Passenger Satisfaction & 0.952 & 0.957 & 0.957 & 0.85 & 0.897 & 0.919 & 0.943 & 0.944 & 0.946 & 0.952 & 0.956 & 0.957  \\
        \hline
      \end{tabular}
\end{table*}

\subsection{Result Analysis} 
The experiment setup described thus far holds in any experiment unless stated otherwise. Our experiments are designed to answer the following three research questions: 
\begin{itemize}[leftmargin=*]
    \item How does the \textit{latent quality} produced by \algo compare to that of the baselines, and to what extent is it superior?
    \item To what extent is the performance affected of each design if the \textit{latent dimension} is varied?
    \item How does the \textit{robustness} of the cached and zero intermediate results differ in \algo against  random client failures?
    \item How efficient is \algo in terms of runtime, network consumption and memory usage compared to the baselines?
\end{itemize}
The questions are answered by conducting three experiments. The first experiment is to evaluate the quality of the latent representation of each design using the classification metrics. The influence of different latent dimensions on the performance is evaluated in the second experiment. To answer the third question, an experiment with different failure probability of each client is devised to compare the cache method for client failure handling with existing zero vector method in the literature.
The amount of guest clients for the first, second and last experiments is fixed to five clients, aligning with established setting in previous research~\cite{zhao2023gtv,Li2023FedSDGFSEA}. The setup of the third experiment consists of one host and two guest clients to adhere to the common two-client setup in VFL ~\cite{Yang2019ParallelDL,Romanini2021PyVerticalAV}. For the third experiment setup, each method runs for 120 epochs in total with batch size of 128.
For the fourth experiment, we set the batch size to 512 and amount of epochs for both training phases to 100.
Early stopping mechanism is disabled in the third and fourth for fair comparison and measurements. 

\subsubsection{\textbf{Latent Quality Experiment}}


Evaluating a design's capability to learn high-quality latent representations offers valuable insights into how effective \algo is able to effectively capture underlying feature correlations compared to the baseline.
The results linked to this experiment are presented in \autoref{fig:latent_eval_results} as histograms. Generally, \codedtext{CT} always outperforms the baseline as expected since it possesses all the data locally. 

Furthermore, \algo clearly performs better than the prior work design (\codedtext{LT}) for all the classification datasets. In terms of accuracy, f1-score and roc-auc \algo can perform up to 22.34\%, 26.12\% and 19.41\% better than \codedtext{LT} respectively. This suggests that \algo is able to capture the underlying feature correlations which leads to the generation of better latent representations of the features. A dataset where \codedtext{LT} was able to perform well and approach evaluation results almost similar to other designs is the \codedtext{intrusion} dataset. This could be attributed by the simplicity of the underlying patterns of the dataset which results in high performance for most of the tested designs. Although \codedtext{TabVFL-LE} has lower accuracy and f1-score compared to \codedtext{LT}, it still managed to be better in terms of ROC-AUC score. This means that \codedtext{TabVFL-LE} is better at discriminating between negative and positive classes (in the multiclass sense) compared to \codedtext{LT}. 

In addition, \codedtext{LT} shows the biggest discrepancies in performance for the \codedtext{forest} and \codedtext{air} datasets compared to \algo. A possible reason is that both datasets have a varying amounts of correlated features at different placements compared to other datasets. 
\codedtext{LT} fails to capture general patterns in such datasets because local pretraining may result in guest clients learning entirely distinct latent representations, introducing potential biases.
This could steer the finetuning model to converge at a suboptimal minima during training, leading to added noise to the latent space, therefore failing to capture common and relevant latent patterns of features scattered among clients. 


\begin{figure*}[h!]
     \centering
     \begin{subfigure}[b]{0.9\linewidth}
            \centering
            \hspace*{1.2cm}\includegraphics[trim=0em 1em 0em 3em, clip, width=1\linewidth]{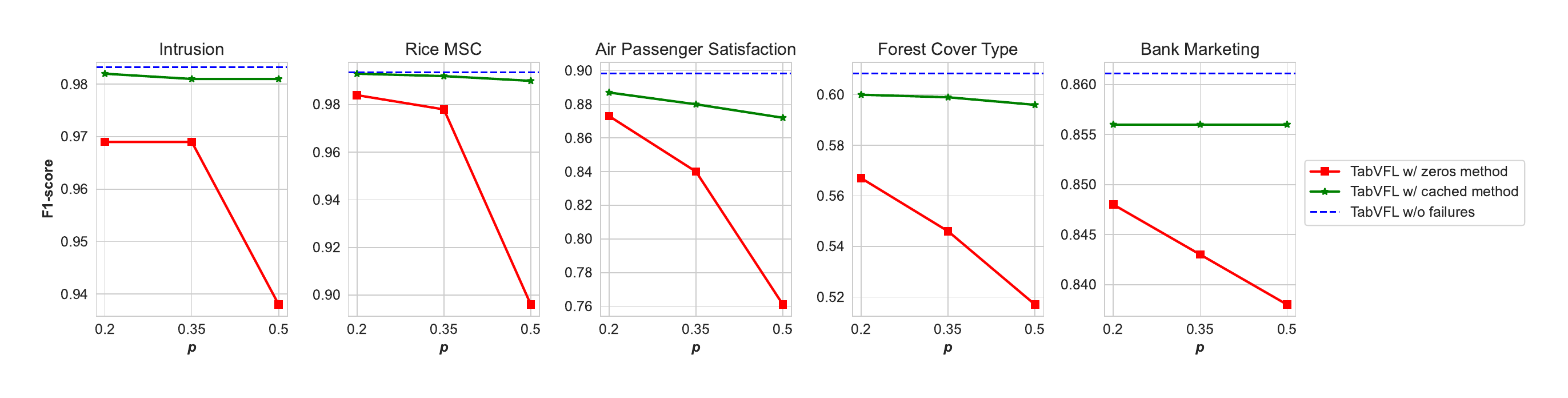}
            \label{fig:client_failure_results_plot}
     \end{subfigure}
     \hfill
     \vspace{-2em}
        \caption{Result over different client failure probabilities  \textit{\textbf{p}} on different datasets.}
        \label{fig:client_failure_results}
\end{figure*}

\begin{table}[t!]
  \centering
  \caption{Client failures experiment improvements of cache method compared to zeros method and maximum performance decrease of both method compared to baseline. All values are presented in percentages. Best results are highlighted in \textbf{bold}.}
  \label{tab:client_failures_improvements}
    \resizebox{\columnwidth}{!}{
      \begin{tabular}{|p{11em}|p{5em}|p{5em}|p{6em}|p{6em}|}
        \hline
        \multirow{2}{*}{Dataset} & \multicolumn{2}{c|}{Cache Method} & Maximum Performance Decrease Zeros Method & Maximum Performance Decrease Cache Method \\
        \cline{2-3}
        & Average Improvement & Maximum Improvement & &\\
        \hline
        Intrusion & 2.3 & 4.37 & -4.78 & \textbf{-0.22}\\
        \hline
        Rice MSC & 3.91 & 9.49 & -10.85 & -0.32\\
        \hline
        Bank Marketing & 1.51 & 2.04 & \textbf{-2.71} & -0.65\\
        \hline
        Forest Cover Type & \textbf{9.15} & \textbf{13.16} & -17.59 & -2.11\\ 
        \hline
        Air Passenger Satisfaction & 6.31 & 12.73 & -18.04 & -3.01\\
        \hline
      \end{tabular}
    }
    \vspace{-1em}
\end{table}

\subsubsection{\textbf{Differing Latent Dimensions Experiment}}

To acquire a comprehensive insight of the performance, we also tested how each design performs for varying latent dimensions. The corresponding results are presented in \autoref{tab:diff-latent-dims}. The setup is identical to the one described in \autoref{sec:exp_details}. We use the latent dimensions 16, 32 and 64 for each design to sufficiently test the effect of small and large values. The \codedtext{CT} design shows somewhat of a stable performance. \codedtext{LT} on the other hand shows an increasing trend in relation with the increasing latent dimension. This is contributed by the larger latent dimension assigned to each client, allowing for an enriched representation of the local data. Furthermore, \codedtext{TabVFL-LE} also demonstrates a positive trend in terms of performance for the majority of the datasets while exhibiting stable performance on others. Moreover, \algo shows similar performance as \codedtext{CT} where the performance is stable on most datasets regardless of the latent dimension. Nevertheless, \codedtext{LT} still underperforms when compared to \algo and the other baselines on most if not all the datasets. The results indicate that \algo is robust against latent dimension changes compared to \codedtext{LT} and can hence be applied in scenarios where stable performance is essential regardless of the latent dimension being used.  

\subsubsection{\textbf{Client Failures Experiment}}

\begin{table}[t!]
  \centering
  \caption{System performance evaluation result. The pretraining and finetuning runtimes on  \codedtext{rice} dataset are presented including the network consumption and memory usage.}
  \label{tab:other_perf_metrics}
  \resizebox{\columnwidth}{!}{
      \begin{tabular}{|p{5em}|p{4em}|p{4em}|p{3.5em}||p{4em}|p{4em}|p{4em}|}
        \hline
        Design & Pretraining & Finetuning & Total Training & Pretraining Network & Finetuning Network & Memory \\
        \hline
        CT & \textbf{151}\,s & \textbf{116}\,s & \textbf{267}\,s & - & - & 
        \textbf{2.75}\,GB
        \\ 
        \hline
        LT & 244\,s & 187\,s & 431\,s & - & \textbf{0.7}\,MB & 
        15.12\,GB
        \\
        \hline
        TabVFL-LE & 252\,s & 204\,s & 456\,s & \textbf{14.7}\,MB & 3.5\,MB & 
        14.73\,GB
        \\
        \hline
        TabVFL & 194\,s & 138\,s & 332\,s & 30.38\,MB & 14.84\,MB & 
        14.73\,GB
        \\
        \hline
      \end{tabular}
    }
\end{table}

In this study, we contrast the caching method for intermediate results (\autoref{subsec:caching_method}) with the zero intermediate results method from \cite{Sun2023RobustAI} in handling client failures. The zero method compensates for missing client results with zero values. Client failure probabilities ranged from 0.2 to 0.5 in 0.15 steps. We also ran a baseline test without client failure handling. Only the f1-score, illustrating client failure impact, is reported and can be seen in \autoref{fig:client_failure_results}.

The dashed blue line represents the baseline without any client failures. The displayed f1-scores are eight-run averages for each probability (\textbf{p}) value. As the client failure probability \textbf{p} increases, the f1-score drops for both methods due to compensating missing results with outdated ones, resulting in suboptimal performance. Yet, cache method shows a more stable and less drastic decline than the zero method. Comparing to zero method, the cache method showed its best average and maximum improvements of 9.15\% and 13.16\%, respectively, on the \codedtext{forest} dataset. In contrast, its worst performance was a 3.01\% decline with the \codedtext{air} dataset, whereas the zeros method dropped by 18.04\% on the same dataset. The minimal performance decline was 0.22\% for the cache method on \codedtext{intrusion} dataset and 2.71\% for the zeros method on \codedtext{bank} dataset. The complete results can be found in \autoref{tab:client_failures_improvements}.

Remarkably, the cache method matched the baseline for the \codedtext{rice} dataset at \textit{p=0.2}, likely due to the dataset's simplicity. Furthermore, caching consistently outperformed the zero method, suggesting higher performance even with client dropouts, by using more realistic intermediate results instead of zeros. Zero vectors might introduce learning biases, producing non-existent patterns. The zero method's declining performance, especially with higher client failure probabilities, contrasts with the cache method's steadier results across datasets. By using the latest intermediate results, the cache method maintains some relevant data representations. However, as seen with the \codedtext{air} dataset, performance might drop, influenced by the dataset complexity and the staleness of cached data.

\subsubsection{\textbf{Framework Efficiency and Resource Consumption Analysis}}
In this section, we measure and display key performance metrics for common VFL systems, including runtime, network consumption, and memory usage for the \algo, \codedtext{LT}, and \codedtext{TabVFL-LE}. For the centralized model, we exclude network consumption as it is not federated. For \codedtext{LT}, as pretraining occurs locally at each guest client, we choose the client with the longest runtime as its representative. No communication means no pretraining network consumption for \codedtext{LT}. The results are presented in \autoref{tab:other_perf_metrics}. Training phase network consumption is gauged using a 4096 batch size for one iteration of each design.

We base our efficiency assessment on the \codedtext{rice} dataset, which has the most data columns in our evaluations, making its results indicative. \algo achieves the fastest runtime, being 29.75\% quicker than \codedtext{LT} and 38.04\% faster than \codedtext{TabVFL-LE}. \codedtext{TabVFL-LE}'s slower pace is due to its aggregation phase requiring iterative summation for each decision step. \codedtext{LT}'s delay stems from the entire TabNet model pretraining at each guest client.

In terms of network consumption during pretraining, \codedtext{TabVFL-LE} uses less than \algo since it transmits outputs (i.e., the latent representation) with a consistent, smaller dimension compared to \algo's output dimension (i.e., the same dimension as input data). For finetuning, \codedtext{LT} is most efficient as each guest uses only a fifth of the total latent dimension, reducing network consumption significantly.

Memory-wise, \codedtext{CT} consumes the least, whereas \codedtext{LT} uses the most. Although \codedtext{TabVFL-LE} guest clients employ TabNet encoders, they have similar memory usage to \algo because \algo's total input dimension equals the combined local input of all \codedtext{TabVFL-LE} guest clients.

\section{Conclusion}
In this paper, we present a novel distributed framework \algo that integrates the state-of-the-art tabular model TabNet to improve latent representation learning on tabular data in the context of vertical federated learning (VFL). To protect against direct data leakage, \algo employs a fully-connected layer to preserve data privacy. This is due to the risk of direct data leakage from integrating TabNet as is in VFL. \algo consolidates intermediate results from all parties in order to learn a single latent respresentation, capturing underlying feature correlations. The framework also addresses the client failures by caching intermediate results. 
Comprehensive experiments across five datasets illustrate remarkable improvement up to 26.12\% on f1-score in latent quality by \algo, surpassing baseline designs. 
\algo demonstrates superior performance in terms of runtime and memory usage, yet encounters reasonable communication overhead when compared to the baselines. 

\bibliographystyle{IEEEtran}
\bibliography{references}

\end{document}